\documentclass[12pt]{article} 
 \usepackage[sectionbib]{natbib}
 \usepackage{array,epsfig,fancyheadings,rotating}
 \usepackage[]{hyperref}  
 \usepackage{sectsty, secdot}
 \sectionfont{\fontsize{12}{14pt plus.8pt minus .6pt}\selectfont}
 
 \subsectionfont{\fontsize{12}{14pt plus.8pt minus .6pt}\selectfont}

 \usepackage{amsmath}
 \usepackage{amssymb}
 \usepackage{amsfonts}
 \usepackage{multirow}
 \usepackage{amsthm}
 \usepackage{dsfont}
 \RequirePackage{mathrsfs,morefloats,booktabs,pdfpages,color,array,hhline,epstopdf,algorithm}

 \newtheorem{thm}{Theorem}
 \newtheorem{lem}{Lemma}

 \theoremstyle{definition}
 \newtheorem{defin}{Definition}

 \newtheorem{rem}{Remark}
\begin{document}


	\title{\bf An improved spectral clustering method for community detection under the degree-corrected stochastic blockmodel}

	\author{
		Huan Qing\thanks{Department of Mathematics,
			China University of Mining and Technology, China. }~
		 and
		Jingli Wang\thanks{Corresponding author. Email: jlwang@nankai.edu.cn. School of Statistics $\&$ Data Science,	Nankai University, China}
		
	}
	\date{}	
	\maketitle
%
	
	\begin{abstract}
		For community detection problem, spectral clustering is a widely used method for detecting clusters in networks. In this paper, we propose an improved spectral clustering (ISC) approach under the degree corrected stochastic block model (DCSBM). ISC is designed based on the k-means clustering algorithm on the weighted leading $K+1$ eigenvectors of a regularized Laplacian matrix where the weights are their corresponding eigenvalues. Theoretical analysis of ISC shows that under mild conditions the ISC yields stable consistent community detection. Numerical results show that ISC outperforms classical spectral clustering methods for community detection on both simulated and eight empirical networks. Especially, ISC provides a significant improvement on two weak signal networks Simmons and Caltech, with error rates of 121/1137 and 96/590, respectively.
	\end{abstract}
	
	\textbf{keyword:}
		Community detection; spectral clustering; weak signal networks; degree-corrected stochastic blockmodel; regularized Laplacian matrix
	


\section{Introduction}
Detecting communities in networks plays a key role in understanding the structure of networks in various areas, including but not limited to computer science, social science, physics, and statistics (\cite{ mcpherson2001birds, duch2005community, fortunato2010community, papadopoulos2012community}). The community detection problem appeals to an ascending number of attention recently (\cite{MN2006,DCSBM,  PJBAC, SCORE}). To solve the community detection problem, substantial approaches, such as \cite{snijders1997estimation,nowicki2001estimation, daudin2008a, PJBAC, rohe2011spectral, amini2013pseudo}, are designed based on the standard framework, the stochastic block model (SBM) (\cite{SBM}), since it is mathematically simple and relatively easy to analyze (\cite{PJBAC}). However, the assumptions of SBM are too restrictive to implement in real networks. It is assumed that the distribution of degrees within the community is Poisson, that is, the nodes within each community have the same expected degrees. Unfortunately, in many natural networks, the degrees follow approximately a power-law distribution (\cite{Kolaczyk2009,Goldenberg2009, SCORE}). The corrected-degree stochastic block model (DCSBM) (\cite{DCSBM}) is developed based on the power-law distribution which allows the degree of nodes varies among different communities. For the community detection problem, a number of methods have been developed based on DCSBM, including some model-based methods and spectral methods. Model-based methods include profile likelihood maximization and modularity maximization as \cite{ DCSBM} and \cite{CMM}. While spectral methods are constructed based on the application of the leading eigenvectors of the adjacency matrix or its variants (see \cite{RSC,SCORE, OCCAM, SLIM}). Compared with spectral methods, model-based methods are pretty time-consuming (\cite{RSC, SCORE, OCCAM}).

A weak signal network can be defined by a network whose adjacency matrix's (or its variants) $(K+1)$-th eigenvalue is close to the $K$-th one in magnitude assuming there are $K$ clusters in the given network (\cite{SCORE+}). However, classical spectral clustering methods such as spectral clustering on ratios-of-eigenvectors (SCORE) (\cite{SCORE}), regularized spectral clustering (RSC) (\cite{RSC}) and overlapping continuous community assignment model (OCCAM) (\cite{OCCAM}) fail to detect two typical  weak signal networks Simmons and Caltech (\cite{SCORE+}). As shown in Table \ref{real2errors}, we find that it is challenging to detect weak signal networks for the recently published symmetrized Laplacian inverse matrix method (SLIM for short) \cite{SLIM}. Though the convexified modularity maximization (CMM) method (\cite{CMM}) and the latent space model based (LSCD) method (\cite{LSCD}) have better performances than SCORE, RSC and OCCAM as shown in the Table 2 of \cite{SCORE+}, these two approaches are time consuming as shown in the Table 4 of \cite{SCORE+}. However, even though \cite{SCORE+} proposed a method called SCORE+ which can be applied to deal with weak signal networks by considering one more eigenvector for clustering, we find that it is challenging and hard to build theoretical framework for it.

In this paper, we aim to seek for a spectral clustering algorithm that has strong statistical performance guarantees under DCSBM, and provides competitive empirical performance when detecting both strong signal networks and weak signal networks.
 We construct a novel Improved Spectral Clustering (ISC) approach under the classical model DCSBM. ISC is designed based on a regularized Laplacian matrix $L_{\delta}$ defined in Section \ref{sec2} and using the production of the leading $(K+1)$ eigenvectors with unit-norm and the leading $(K+1)$ eigenvalues for clustering. Some theoretical results are established to ensure a stable performance of the proposed method. As we show in numerical studies, our approach ISC has satisfactory performances in simulated networks and empirical examples, especially, for two real-world weak signal networks Simmons and Caltech (introduced in Section \ref{secmotivation}). In all, our approach is comparable to or improves upon the published state-of-the-art methods in the literature.

The details of motivation are described in section \ref{secmotivation}. In Section \ref{sec2}, we propose the ISC approach after introducing the classical DCSBM model. Section \ref{sec3} presents the theoretical framework of ISC where we show that the ISC yields stable consistent community detection under mild conditions. Section \ref{sec4} investigates the performance of the ISC via comparing with five spectral clustering methods on both numerical networks and eight empirical datasets. Section \ref{sec5} concludes.

\section{Motivation}\label{secmotivation}
Before introducing our algorithm, we briefly describe two empirical networks Simmons and Caltech which inspire us to design our algorithm.

The Simmons network induced by nodes with graduation year between 2006 and 2009 has a largest connected component with 1137 nodes. \cite{traud2011comparing} found that there exists community structure in the Simmons network based on the observation that there is a strong correlation with the graduation year such that students enrolled by the college in the same year are more likely to be friends (i.e., node in Simmons network denotes student, and edge denotes the friendship). Naturally, the Simmons network contains four communities such that students in the same year are in the same community, which is applied as the ground truth to test the performances of community detection approaches.

The Caltech network contains a largest component with 590 nodes and \cite{traud2011comparing} found that the community structure in the Caltech network is highly correlated with the eight dorms that a student is from such that students in the same dorm are more likely to be friends (i.e., node in Caltech network denotes student, and edge denotes the friendship). Therefore, the Caltech network contains eight communities such that students in the same dorm are in the same community, which is applied as the ground truth in this paper.

Simmons (as well as Caltech) is regarded as a weak signal network since the leading $(K+1)$-th eigenvalue of the adjacency matrix or its variants are close to the leading $K$-th eigenvalue assuming it has $K$ clusters. \cite{SCORE+} found that the leading $(K+1)$-th eigenvector of the adjacency matrix or its variants may also contain information about nodes labels for weak signal networks. The intuition behind our method is that published spectral methods such as SCORE (\cite{SCORE}), RSC (\cite{RSC}) and OCCAM (\cite{OCCAM}) fail to detect Simmons and Caltech as shown in \cite{SCORE+}. The error rates (defined in \ref{DefinError}) of our ISC and other four spectral methods SCORE, OCCAM, RSC, and SLIM on the two weak signal networks are recorded by Table \ref{real2errors}, from which we can find that SCORE, OCCAM, RSC, and SLIM perform poor with high error rates while our ISC has satisfactory performances on the two datasets Simmons and Caltech.
\begin{table}[t!]
	\centering
	\caption{Error rates on the two weak signal networks Simmons and Caltech.}
	\label{real2errors}\par
	\resizebox{\linewidth}{!}{
	\begin{tabular}{cccccc}
	\hline
		&ISC&SCORE&OCCAM&RSC&SLIM\\
		\hline
		Simmons&\textbf{121/1137}&268/1137&268/1137&244/1137&275/1137\\
		Caltech&\textbf{96/590}&180/590&192/590&170/590&150/590\\
		\hline
	\end{tabular}}
\end{table}

\section{Problem setup and the ISC algorithm}\label{sec2}
In this section, we set up the community detection problem under DCSBM, and then introduce our algorithm ISC.

The following notations will be used throughout the paper: $\|\cdot\|_{F}$ for a matrix denotes the Frobenius norm. $\|\cdot\|$ for a matrix denotes the spectral norm. $\|\cdot\|$ for a vector denotes the $l_{2}$-norm. For convenience, when we say ``leading eigenvectors'' or ``leading eigenvalues'', we are comparing the \emph{magnitudes} of the eigenvalues. For any matrix or vector $x$, $x'$ denotes the transpose of $x$.

Consider an undirected, no-loops, and unweighted connected network $\mathcal{N}$ with $n$ nodes. Let $A$ be an adjacency matrix of network $\mathcal{N}$ such that $A_{ij}=1$ if there is an edge between node $i$ and $j$, $A_{ij}=0$ otherwise, for $i,j = 1, \dots, n$ ($A$ is symmetric with diagonal entries being zeros). We assume that there exist $K$ perceivable non-overlapping clusters
\begin{align}\label{definV}
V^{(1)}, V^{(2)}, \ldots, V^{(K)},
\end{align}
and each node belongs to exactly only one cluster, and $K$ is assumed to be known in this paper. Let $\ell$ be an $n\times 1$ vector such that $\ell(i)$ takes values from $\{1, 2, \ldots, K\}$ and $\ell(i)$ is the cluster label for node $i$, $i= 1, \cdots, n$. $\ell$ is unknown and our goal is to estimate $\ell$ with given $(A, K)$ .

\subsection{The degree-corrected stochastic block model}
In this paper, we consider the degree-corrected stochastic block model (DCSBM) (\cite{DCSBM}) which is a generalization of the classic stochastic block model (SBM) (\cite{SBM}). Under DCSBM, a set of tuning parameters are used to control the node degrees. $\theta$ is the $n\times 1$ degree vector and $\theta_{i}>0$ is the $i$-th element of $\theta$. Let $P$ be a $K\times K$ matrix such that
\begin{align}\label{definP}
P\mathrm{~is~symmetric,~nonsingular, nonnegative~and~irreducible}.
\end{align}
Generally, the elements of $P$ denote the probability of generating an edge between distinct nodes, therefore the entries of $P$ are always assumed to be in the interval $[0, 1]$. In this paper, we call $P$ as the mixing matrix for convenience. Then under DCSBM, the probability of an edge between node $i$ and node $j$ is
\begin{align}\label{definprobability}
Pr(A_{ij}=1)=\theta_{i}\theta_{j}P_{g_{i}g_{j}},
\end{align}
where $\theta_{i}\theta_{j}P_{g_{i}g_{j}}\in [0,1]$ and $g_{i}$ denotes the cluster that node $i$ belongs to. Note that if $g_{i}=g_{j}$ for any two distinct nodes $i,j$ such  that $\theta_{i}=\theta_{j}$, the DCSBM model degenerates to SBM.
Then the degree heterogeneity matrix $\Theta \in \mathcal{R}^{n\times n}$ and the $n\times K$ membership matrix $Z\in \mathcal{R}^{n\times K}$ can be presented as:
\begin{align}\label{definZ}
\Theta=\mathrm{diag}(\theta_{1},\theta_{2},\ldots,
\theta_{n}),~Z_{ik}=\mathds{1}_{\{g_{i}=k\}} \mathrm{~for~}1\leq i\leq n, 1\leq k\leq K.
\end{align}
\begin{defin}
Model (\ref{definV})-(\ref{definZ}) constitute the so-called degree-corrected stochastic block model (DCSBM), and denote it by $DCSBM(n, P, \Theta, Z)$ for convenience.
\end{defin}
\subsection{The algorithm: ISC}

Under $DCSBM(n, P, \Theta, Z)$, the details of ISC method proceed as follows:

\textbf{ISC}. Input:  $A, K$, and a ridge regularizer $\delta>0$. Output: community labels for all nodes.

\textbf{Step 1}: Obtain the graph Laplacian with ridge regularization by
\begin{align*}
  L_{\delta}=(D+\delta dI)^{-1/2}A(D+\delta d I)^{-1/2},
\end{align*}
where $d=\frac{d_{\mathrm{max}}+d_{\mathrm{min}}}{2}, d_{\mathrm{max}}=\mathrm{max}_{1\leq i\leq n}d_{i}, d_{\mathrm{min}}=\mathrm{min}_{1\leq i\leq n}d_{i}$ (where $d_{i}$ is the degree of node $i$, i.e., $d_{i}=\sum_{j=1}^{n}A_{ij}$ for $1\leq i\leq n$), and $I$ is the $n\times n$ identity matrix. Note that the ratio of the largest diagonal entry of $D+\delta d I$ and the smallest one is in $[\frac{d_{\mathrm{min}}}{d_{\mathrm{max}}}, \frac{d_{\mathrm{max}}}{d_{\mathrm{min}}}]$. Conventional choices for $\delta$ are 0.05 and 0.10.

\textbf{Step 2}:  Compute the leading $\textbf{K+1}$ eigenvalues and eigenvectors with unit norm of $L_{\delta}$, and then calculate  the weighted eigenvectors matrix:
\begin{align*}
  \hat{X}=[\hat{\eta}_{1},\hat{\eta}_{2}, \ldots, \hat{\eta}_{K}, \hat{\eta}_{K+1}]\cdot \mathrm{diag}(\hat{\lambda}_{1}, \hat{\lambda}_{2}, \ldots, \hat{\lambda}_{K}, \hat{\lambda}_{K+1}),
\end{align*}
where $\hat{\lambda}_{i}$ is the $i$-th leading eigenvalue of $L_{\delta}$, and $\hat{\eta}_{i}$ is the respective eigenvector with unit-norm, for $i = 1, \cdots, K+1$.

\textbf{Step 3}:  Normalizing each row of $\hat{X}$  to have unit length, and denote by $\hat{X}^{*}$,
$$\hat{X}^{*}_{ij}=\hat{X}_{ij}/(\sum_{j=1}^{K+1}\hat{X}_{ij}^{2})^{1/2}, i = 1, \dots, n, j=1, \dots, K+1.$$

\textbf{Step 4}: For clustering, apply k-means method to $\hat{X}^{*}$ with assumption that there are $K$ clusters.

Compared with tradition spectral clustering methods such as SCORE, RSC and OCCAM, our ISC is quite different from them. The differences can be stated in the following four aspects: (a) ISC applies  a graph Laplacian with ridge regularization $L_{\delta}$ instead of the graph Laplacian $L_{\tau}$ in RSC. (b) ISC uses eigenvalues to re-weight the columns of $\hat{X}$, i.e., the $\hat{X}$ in ISC is computed as the production of eigenvectors and eigenvalues, instead of simply using the eigenvectors as in SCORE and RSC. Meanwhile, we apply leading eigenvalues with power 1 instead of power 0.5 in OCCAM, which ensures that our ISC can detect dis-associative networks while OCCAM can not. (c) ISC always selects the leading $(K+1)$ eigenvectors and eigenvalues for clustering while other spectral clustering approaches only apply the leading $K$ eigenvectors for clustering, such as SCORE and RSC. This main characteristic enables the ISC to deal with weak signal networks. (d) RSC takes the average degree as default regularizer $\tau$, and SCORE is sensitive to the choice of threshold, while our ISC is insensitive to the choice of $\delta$ as long as it is slightly larger than zero (see Section \ref{sec4} for details). All these four aspects are significant in supporting the performances of our ISC. The numerical comparison of these methods is demonstrated in Section \ref{sec4}.
\section{Theoretical Results}\label{sec3}
\subsection{Population analysis of the ISC}
This section presents the population analysis of ISC method to show that it returns perfect clusters in the ideal case.

 Under $DCSBM(n, P, \Theta, Z)$, denote the expectation matrix of the adjacency matrix $A$ as $\Omega =\mathop{{}\mathbb{E}}[A]$ such that $\Omega_{ij}=Pr(A_{ij}=1)=\theta_{i}\theta_{j}P_{g_{i}g_{j}}$. Following \cite{SCORE}, \cite{DCSBM} and \cite{RSC}, matrix $\Omega$ can be expressed as
\begin{align*}
  \Omega=\Theta ZPZ'\Theta.
\end{align*}
Then the population Laplacian can be written as
\begin{align*}
\mathscr{L}=\mathscr{D}^{-1/2}\Omega \mathscr{D}^{-1/2},
\end{align*}
where $\mathscr{D}$ is a diagonal matrix which contains the expected node degrees, $\mathscr{D}_{ii}=\sum_{j}^{n}\Omega_{ij}$. And the regularized Laplacian  $\mathscr{L}_{\delta}$ in $\mathcal{R}^{n\times n}$ can be presented as
\begin{align*}
\mathscr{L}_{\delta}=\mathscr{D}_{\delta}^{-1/2}\Omega\mathscr{D}_{\delta}^{-1/2},
\end{align*}
where $\mathscr{D}_{\delta}=\mathscr{D}+\delta d I$, $\delta$ and $d$ are tuning parameters.

We find that the regularized Laplacian matrix has an explicit form which is a product of some matrices. For convenience, we introduce some notations. Let $Q$ be a $K\times n$ matrix with $Q=PZ'\Theta$, and let $D_{P}$ be a
$K\times K$ diagonal matrix with $D_{P}(i,i)=\sum_{j=1}^{n}Q_{ij}, i=1,2,\ldots,K$. Define $\theta^{\delta}$ as an $n\times 1$ vector whose $i$-th entry $\theta^{\delta}_{i}$ is $\theta_{i}\frac{\mathscr{D}_{ii}}{\mathscr{D}_{ii}+\delta d}$. And let $\Theta_{\delta}$ be a diagonal matrix whose $ii$'th entry is $\theta^{\delta}_{i}$ for $1\leq i\leq n$. Define a $K\times K$ symmetric matrix $\tilde{P}$ as $\tilde{P}=D_{P}^{-1/2}PD_{P}^{-1/2}$.
Based on these notations, Lemma \ref{formofL} gives the explicit form for $\mathscr{L}_{\delta}$.
\begin{lem}\label{formofL}
(\emph{Explicit form for} $\mathscr{L}_{\delta}$) Under $DCSBM(n, P, \Theta, Z)$, $\mathscr{L}_{\delta}$ can be written as
\begin{align*}
  \mathscr{L}_{\delta}=\mathscr{D}_{\delta}^{-1/2}\Omega \mathscr{D}_{\delta}^{-1/2}=\Theta_{\delta}^{1/2}Z\tilde{P}Z'\Theta_{\delta}^{1/2}.
\end{align*}
\end{lem}
Note that Lemma \ref{formofL} is equivalent to the Lemma 3.2 in \cite{RSC}, so we omit the proof of it here.

To express the eigenvectors of $\mathscr{L}_{\delta}$, we rewrite $\mathscr{L}_{\delta}$ in the following form:
\begin{align*}
\mathscr{L}_{\delta}=\|\tilde{\theta}\|^{2}\tilde{\Gamma}\tilde{D}\tilde{P}\tilde{D}\tilde{\Gamma}',
\end{align*}
where $\tilde{\theta} = (\tilde{\theta}_{1}, \dots, \tilde{\theta}_{n})' = ( \sqrt{\theta_{1}^{\delta}}, \dots, \sqrt{\theta_{n}^{\delta}} )',$ $\tilde{\theta}^{(k)}$ is an $n\times 1$  vector with elements $\tilde{\theta}^{(k)}_{i} = \tilde{\theta}_{i}\mathds{1}_{(g_{i}=k)}$ for $1\leq i\leq n, 1\leq k\leq K$,  $\tilde{D}$ is a $K\times K$ diagonal matrix of the \emph{overall degree intensities} such that $\tilde{D}(k,k)=\|\tilde{\theta}^{(k)}\|/\|\tilde{\theta}\|$, and $\tilde{\Gamma}$ is an $n\times K$ matrix  such that
\begin{align*}
\tilde{\Gamma}=[\frac{\tilde{\theta}^{(1)}}{\|\tilde{\theta}^{(1)}\|} ~ \frac{\tilde{\theta}^{(2)}}{\|\tilde{\theta}^{(2)}\|} ~\ldots~\frac{\tilde{\theta}^{(K)}}{\|\tilde{\theta}^{(K)}\|}].
\end{align*}
Note that $\tilde{\Gamma}'\tilde{\Gamma}=I$, where $I$ is the $K\times K$ identity matrix.


By the definition (\ref{definP}) of $P$ and basic knowledge of algebra, we know that the rank of $\mathscr{L}_{\delta}$ is $K$ when there are $K$ clusters, therefore $\mathscr{L}_{\delta}$ has $K$ nonzero eigenvalues. Lemma \ref{eigenvectors} gives the expressions of the leading $K$ eigenvectors of $\mathscr{L}_{\delta}$.
\begin{lem}\label{eigenvectors}
Under $DCSBM(n, P, \Theta, Z)$, suppose all eigenvalues of $\tilde{D}\tilde{P}\tilde{D}$ are simple. Let $\lambda_{1}/\|\tilde{\theta}\|^{2}, \lambda_{2}/\|\tilde{\theta}\|^{2},\ldots, \lambda_{K}/\|\tilde{\theta}\|^{2}$ be such eigenvalues, arranged in the descending order of the magnitudes, and let $a_{1}, a_{2}, \ldots, a_{K}$ be the associated (unit-norm) eigenvectors. Then the $K$ nonzero eigenvalues of $\mathscr{L}_{\delta}$ are $\lambda_{1}, \lambda_{2}, \ldots, \lambda_{K}$, and the associated (unit-norm) eigenvectors are
\begin{align*}
  \eta_{k}=\sum_{i=1}^{K}[a_{k}(i)/\|\tilde{\theta}^{(i)}\|]\cdot \tilde{\theta}^{(i)}, k=1,2,\ldots, K.
\end{align*}
\end{lem}

Based on the regularized Laplacian $\mathscr{L}_{\delta}$ and its eigenvalues and eigenvectors, we can write down the Ideal ISC algorithm as follows:

 \textbf{Ideal ISC}. Input: $\Omega$. Output: $\ell$.

 \textbf{Step 1:} Obtain $\mathscr{L}_{\delta}$.

 \textbf{Step 2:} Obtain $X=[\eta_{1},\eta_{2}, \ldots, \eta_{K}, \eta_{K+1}]\cdot \mathrm{diag}(\lambda_{1}, \lambda_{2}, \ldots, \lambda_{K}, \lambda_{K+1})$. Since $\mathscr{L}_{\delta}$ only has $K$ nonzero eigenvalues (i.e., $\lambda_{i}\neq 0$ for $1\leq i\leq K$ and $\lambda_{K+1}=0$), we have $X=[\lambda_{1}\eta_{1},\lambda_{2}\eta_{2}, \ldots, \lambda_{K}\eta_{K}, \lambda_{K+1}\eta_{K+1}]
 =[\lambda_{1}\eta_{1},\lambda_{2}\eta_{2}, \ldots, \lambda_{K}\eta_{K}, 0].$

 \textbf{Step 3:} Obtain $X^{*}$, the row-normalized version of $X$.

 \textbf{Step 4:} Apply k-means to $X^{*}$,  assuming there are $K$ clusters.

Note that the Ideal ISC algorithm is obtained by applying $\Omega$ to replace the input the adjacency matrix $A$ in the ISC algorithm, that's why we call it as the Ideal ISC algorithm.
Lemma \ref{population} confirms that the Ideal ISC algorithm can return the perfect clustering under $DCSBM(n, P, \Theta, Z)$.
\begin{lem}\label{population}
Under $DCSBM(n, P, \Theta, Z)$, $X$ has $K$ distinct rows, and for any two distinct nodes $i, j$, if $g_{i}=g_{j}$, the $j$-th row of $X$ equals the $i$-th row of it.
\end{lem}
Applying $K$-means to $X$ leads to the true community labels of
each node. The above population analysis for ISC method presents a direct understanding of why the ISC algorithm works and guarantees that ISC method returns perfect clustering results under the ideal case.
\subsection{Characterization of the matrix $X^{*}$}
We wish to bound $\|\hat{X}^{*}-X^{*}\|_{F}$. To do so, first we need to find the bound of $\|\hat{X}-X\|_{F}$ based on the bound of the difference between eigenvalues of $L_{\delta}$ and $\mathscr{L}_{\delta}$. 
Then, we give Lemma \ref{boundeigenvalue} to bound the difference between eigenvalues of $L_{\delta}$ and $\mathscr{L}_{\delta}$.
\begin{lem}\label{boundeigenvalue}
  Under $DCSBM(n, P, \Theta, Z)$, if $d\delta+\mathrm{min}_{i}\mathscr{D}_{ii}>3\mathrm{log}(4n/\epsilon)$, with probability at least $1-\epsilon$,
\begin{align*}
\underset{1\leq k\leq K}{\mathrm{max}}|\hat{\lambda}_{k}-\lambda_{k}|\leq 4\sqrt{\frac{3\mathrm{log}(4n/\epsilon)}{d\delta +\mathrm{min}_{i}\mathscr{D}_{ii}}} \mathrm{~and~} \sum_{i=1}^{K}|\hat{\lambda}_{i}-\lambda_{i}|^{2}\leq \frac{48K\mathrm{log}(4n/\epsilon)}{d\delta+\mathrm{min}_{i}\mathscr{D}_{ii}}.
\end{align*}
\end{lem}
Let $\hat{X}_{i}, X_{i}$ denote the $i$-th row of $\hat{X}$ and $X$, respectively. Next, we apply results of Lemma \ref{boundeigenvalue} to obtain the bound for $\|\hat{X}^{*}-X^{*}\|_{F}$, which is given in the following lemma.
\begin{lem}\label{boundstar}
Under $DCSBM(n, P, \Theta, Z)$, define $m=\mathrm{min}_{i}\{\mathrm{min}\{\|\hat{X}_{i}\|, \|X_{i}\|\}\}$ as the length of the shortest row in $\hat{X}$ and $X$. Then, for any $\epsilon >0$ and sufficiently large $n$,  assume that
\begin{align*}
(a)~~ \big(\frac{\mathrm{log}(4n/\epsilon)}{d\delta+\mathrm{min}_{i}\mathscr{D}_{ii}}\big)^{1/2}\leq \frac{\lambda_{K}}{8\sqrt{3}},~~
(b)~~d\delta+\mathrm{min}_{i}\mathscr{D}_{ii}>3\mathrm{log}(4n/\epsilon),
\end{align*}
with probability at least $1-\epsilon$, the following hold
\begin{align*}
&\|\hat{X}-X\|_{F}\leq\big(\frac{48K\mathrm{log}(4n/\epsilon)}{d\delta +\mathrm{min}_{i}\mathscr{D}_{ii}}+\hat{\lambda}_{K+1}\big)^{1/2}+\frac{32K}{\lambda_{K}}\big(\frac{3\mathrm{log}(4N/\epsilon)}{d\delta+\mathrm{min}_{i}\mathscr{D}_{ii}}\big)^{1/2},\\
&\|\hat{X}^{*}-X^{*}\|_{F}\leq \frac{1}{m}\big(\frac{48K\mathrm{log}(4n/\epsilon)}{d\delta +\mathrm{min}_{i}\mathscr{D}_{ii}}+\hat{\lambda}_{K+1}\big)^{1/2}+\frac{32K}{m\lambda_{K}}\big(\frac{3\mathrm{log}(4N/\epsilon)}{d\delta+\mathrm{min}_{i}\mathscr{D}_{ii}}\big)^{1/2}.
\end{align*}
\end{lem}
\subsection{Bound of Hamming error rate of ISC}
This section bounds the Hamming error rate (\cite{SCORE}) of ISC under DCSBM to show that ISC yields stable consistent community detection.

 The Hamming error rate of ISC is defined as:
 \begin{align*}
   \mathrm{Hamm}_{n}(\hat{\ell},\ell)=\underset{\pi\in S_{K}}{\mathrm{min~}}H_{p}(\hat{\ell},\pi(\ell))/n,
 \end{align*}
 where $S_{K}=\{\pi:\pi \mathrm{~is~a~permutation~of~the~set~}\{1,2,\ldots, K\}\}$ \footnote{ Due to the fact that the clustering errors should not
depend on how we tag each of the K communities, that's why we need to consider permutation to measure the clustering errors of ISC here.}, $\pi(\ell)$ is the $n\times 1$ vector such that $\pi(\ell)(i)=\pi(\ell(i))$ for $1\leq i\leq n$,  and $H_{p}(\hat{\ell},\ell)$ is the expected number of mismatched labels defined as below
 \begin{align*}
   H_{p}(\hat{\ell},\ell)=\sum_{i=1}^{n}P(\hat{\ell}(i)\neq \ell(i)).
 \end{align*}
 The following theorem is the main theoretical result of ISC, which bounds the Hamming error of ISC algorithm under mild conditions and shows that the ISC method yields stable consistent community detection if we assume that the adjacency matrix $A$ are generated from the DCSBM model.
\begin{thm}\label{mainISC}
Set $err_{n}=4\big[\frac{1}{m}\sqrt{\frac{48K\mathrm{log}(4n/\epsilon)}{d\delta +\mathrm{min}_{i}\mathscr{D}_{ii}}+\hat{\lambda}^2_{K+1}}+\frac{32K}{m\lambda_{K}}\sqrt{\frac{3\mathrm{log}(4N/\epsilon)}{d\delta+\mathrm{min}_{i}\mathscr{D}_{ii}}}\big]^{2}$,
 under $DCSBM(n, P, \Theta, Z)$ and the same assumptions as in Lemma \ref{boundstar} hold, suppose as $n\rightarrow \infty$, we have
\begin{align*}
  err_{n}/\mathrm{min~}\{n_{1}, n_{2}, \ldots, n_{K}\}\rightarrow 0,
\end{align*}
where $n_{k}$ is the size of the $k$-th community for $1\leq k\leq K$.
For the estimated label vector $\hat{\ell}$ by ISC,  with probability at least $1-\epsilon$, we have
\begin{align*}
  \mathrm{Hamm}_{n}(\hat{\ell},\ell) \leq err_{n}/n.
\end{align*}
\end{thm}
Note that by the assumption (a) in Theorem \ref{boundstar}, we can zoom $\mathrm{Hamm}_{n}(\hat{\ell},\ell)$ as
\begin{align}\label{HammStrong}
\mathrm{Hamm}_{n}(\hat{\ell},\ell)\leq\frac{(\sqrt{K\lambda_{K}^{2}+4\hat{\lambda}^{2}_{K+1}}+8K)^{2}}{nm^{2}},
\end{align}
which tells us that 1) with the increasing of $n$, while keeping other parameters fixed, the Hamming error rates of ISC decreases to zero; 2) a larger $K$ suggests a larger error bound, which means that it becomes harder to detect communities for ISC when $K$ increases; 3) for weak signal networks, since $\hat{\lambda}_{K+1}$ is close to $\hat{\lambda}_{K}$ in magnitude, and we also know that $\lambda_{K}$ is close to $\hat{\lambda}_{K}$ by Lemma \ref{boundeigenvalue}, we can conclude that $\hat{\lambda}_{K+1}$ is close to $\lambda_{K}$ in some sense, therefore, for weak signal networks, we can even roughly simplify the bound of $\mathrm{Hamm}_{n}(\hat{\ell},\ell)$ by
\begin{align}\label{HammWeak}
\mathrm{Hamm}_{n}(\hat{\ell},\ell)\leq\frac{(|\hat{\lambda}_{K+1}|\sqrt{K+4}+8K)^{2}}{nm^{2}}.
\end{align}
As for strong signal networks, $\hat{\lambda}_{K}$ is much larger than  $\hat{\lambda}_{K+1}$ in magnitude, by Lemma \ref{boundeigenvalue},  $4\hat{\lambda}^{2}_{K+1}$  is also smaller than $K\lambda^{2}_{K}$ for $K\geq 2$, hence the bound of the Hamming error rate for strong signal networks mainly depends on the $K$-th leading eigenvalue.

Combining (\ref{HammStrong}) and (\ref{HammWeak}), we can find that our ISC can detect both strong signal networks and weak signal networks. The main reason is that its theoretical bound of clustering error depends on the leading $K$-th eigenvalue of $L_{\delta}$ for strong signal networks and the bound relies on the leading $(K+1)$-th eigenvalue of $L_{\delta}$ when dealing with weak signal networks.
\section{Numerical Results}\label{sec4}
We compare ISC with a few recent methods: SCORE, OCCAM, RSC, SCORE+ and SLIM via synthetic data and eight real-world networks. Unless specified, we set the regularizer $\delta=0.1$ for ISC in this section. For each procedure, the clustering error rate is measured by
\begin{align}\label{DefinError}
  \mathrm{min}_{\{\pi: \mathrm{permutation~over~}\{1,2,\ldots, K\}\}}\frac{1}{n}\sum_{i=1}^{n}1\{\pi(\hat{\ell}(i))\neq \ell(i)\}.
\end{align}
where $\ell(i)$ and $\hat{\ell}(i)$ are the true and estimated labels of node $i$.
\subsection{Synthetic data experiment}
In this subsection, we use two simulated experiments to investigate the performances of these approaches.

\textbf{\texttt{Experiment 1}}. We  investigate performances of these approaches under SBM when $K=2$ and 3 by increasing $n$. Set $n\in \{40, 80, 120, \ldots, 400\}$. For each fixed $n$, we record the mean of the error rates of 50 repetitions. Meanwhile, to check that whether we generate weak signal networks during simulation, we also (including Experiment 2) record the mean of the quantity $1-|\frac{\lambda_{K+1}}{\lambda_{K}}|$ where $\lambda_{k}$ $(k=K, K+1)$ denotes the $k$-th leading eigenvalue of $A$ and $L_{\delta}$.

\emph{Experiment 1(a)}. Let $K=2$ in this sub-experiment. Generate $\ell$ by setting each node belonging to one of the clusters with equal probability (i.e., $\ell(i)-1 \overset{\mathrm{i.i.d.}}{\sim}\mathrm{Bernoulli}(1/2)$). Set the mixing matrix $P_{1(a)}$ as
 \[
P_{1(a)}
=
\begin{bmatrix}
    0.9&0.5\\
    0.5&0.9\\
\end{bmatrix}.
\]
Generate $\theta$ as $\theta(i)=0.2$ for $g_{i}=1$, $\theta(i)=0.6$ for $g_{i}=2$.

\emph{Experiment 1(b)}. Let $K=3$ in this sub-experiment.  $\ell$ is generated same as in \emph{Experiment 1(a)}. Set the mixing matrix $P_{1(b)}$ as
 \[
P_{1(b)}
=
\begin{bmatrix}
    0.9&0.5&0.5\\
    0.5&0.9&0.5\\
    0.5&0.5&0.9\\
\end{bmatrix}.
\]
Generate $\theta$ as $\theta(i)=0.2$ for $g_{i}=1$, $\theta(i)=0.6$ for $g_{i}=2$, and $\theta(i)=0.8$ for $g_{i}=3$.

\begin{figure}[H]
\includegraphics[width=12cm,height=4cm]{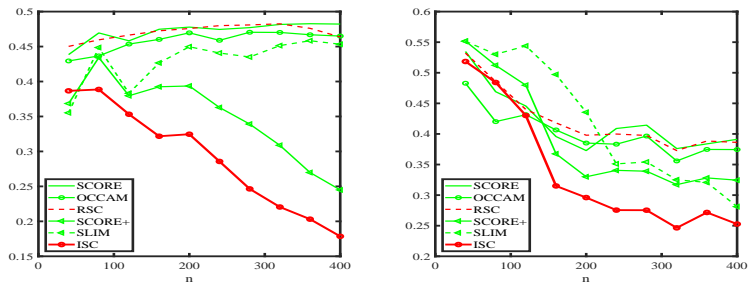}
\caption{Numerical results of Experiment 1. Left panel: Experiment 1(a). Right panel: Experiment 1(b). y-axis: error rates.}\label{Ex1}
\end{figure}

\begin{figure}[H]
\includegraphics[width=12cm,height=4cm]{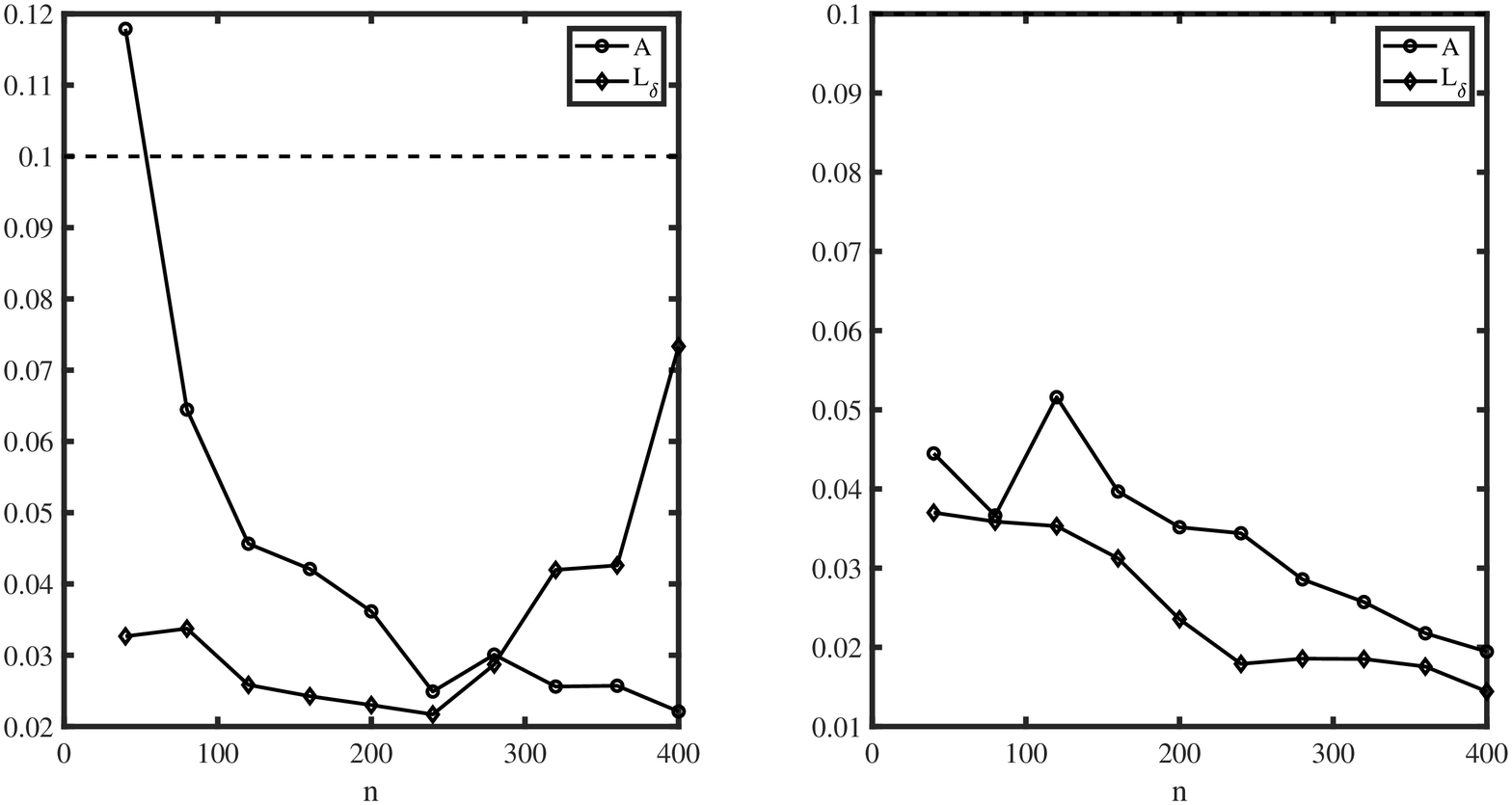}
\caption{Numerical results of Experiment 1. Left panel: Experiment 1(a). Right panel: Experiment 1(b). y-axis: $1-|\frac{\lambda_{K+1}}{\lambda_{K}}|$. The black circle line denotes that the two eigenvalues are from $A$ in the quantity $1-|\frac{\lambda_{K+1}}{\lambda_{K}}|$, while the black diamond line denotes that the two eigenvalues are from $L_{\delta}$ (we use the same labels in Experiment 2).}\label{eigEx1}
\end{figure}

The numerical results of Experiment 1 are shown in Figure \ref{Ex1} from which we can find that 1) ISC outperforms the other five procedures obviously in Experiment 1(a) and 1(b). 2) error rates of ISC and SCORE+ decrease as $n$ increases, while the other four approaches perform unsatisfactory even when $n$ increases in Experiment 1(a). This phenomenon occurs because even the sample size $n$ is increasing to 400, it is still too small for SCORE, OCCAM, RSC and SLIM. Furthermore, by Figure \ref{eigEx1}, we can find that the quantity $1-|\frac{\lambda_{K+1}}{\lambda_{K}}|$ is almost always smaller than 0.1 when the eigenvalue $\lambda$ is from the adjacency matrix $A$ and $1-|\frac{\lambda_{K+1}}{\lambda_{K}}|$ is always smaller than 0.1 when $\lambda$ is from the regularized Laplacian $L_{\delta}$ in this experiment, which indicate that $\lambda_{K}$ and $\lambda_{K+1}$ are close enough. By the definition of weak signal networks, the networks  generated in this experiment are weak signal networks\footnote{As far as we know, there are two real-world weak signal networks Simmons and Caltech (\cite{SCORE+}), which suggests that generating weak signal networks numerically is significant since it can help us to test whether our newly designed methods can deal with weak signal networks, not only depending on the performances of the two real-world weak signal networks Simmons and Caltech.}. In all, ISC outperforms other methods when the network is weak signal.

\textbf{\texttt{Experiment 2}}. In this experiment, we study how decreasing the difference of
 $\theta$ between different communities, the ratios of the size between different communities, and the ratios of diagonal and off-diagonal entries of the mixing matrix impact the performance of our ISC approach when $n=400$ and $K=2$.

\emph{Experiment 2(a)}. We study  the influence of decreasing the difference of $\theta$ between different communities under SBM in this sub-experiment. Set $a_{0}$ in $\{0.05, 0.1, 0.15, \ldots, 0.45, 0.5\}$. We generate $\ell$  by setting each node belonging to one of the clusters with equal probability. And set the mixing matrix $P_{2(a)}$ as
 \[\renewcommand{\arraystretch}{0.75}
P_{2(a)}
=
\begin{bmatrix}
    0.9&0.5\\
    0.5&0.9\\
\end{bmatrix}.
\]
Generate $\theta$ as $\theta(i)=1-a_{0}$ if $g_{i}=1$ ,$\theta(i)=a_{0}$ for $g_{i}=2$. Note that for each $a_{0}$, $\theta(i)$ is a fixed number for all nodes in the same community and hence this is a SBM case. For each $a_{0}$ (as well as $b_{0}$ and $c_{0}$ in the following experiments), we record the mean of clustering error rates of 50 sampled networks.

\emph{Experiment 2(b)}. We study how the proportion between the diagonals and off-diagonals of $P$ affects the performance of these methods under SBM in this sub-experiment. Set the proportion $b_{0}$  in $\{1/20, 2/20, 3/20,\ldots, 10/20\}$. We generate $\ell$  by setting each node belonging to one of the clusters with equal probability. Set $\theta$ as $\theta(i)=0.2$ if $g_{i}=1$ and $\theta(i)=0.6$ otherwise. The mixing matrix $P_{2(b)}$ is set as below:
 \[\renewcommand{\arraystretch}{0.75}
P_{2(b)}
=
\begin{bmatrix}
   0.5& b_{0} \\
    b_{0} & 0.5\\
\end{bmatrix}.
\]

\emph{Experiment 2(c)}. All parameters are same as Experiment 2(b), except that we set $g_{i}=1$ for $1\leq i\leq 100$, $g_{i}=2$ for $101\leq i\leq n$, and $\theta_{i}=0.5+0.5(i/n)^{2}$ for $1\leq i\leq n$ (i.e., Experiment 2(c) is the DCSBM case).

\emph{Experiment 2(d)}. We study how the proportion between the size of clusters influences the performance of these methods under SBM in this sub-experiment. We set the proportion $c_{0}$ in $\{1,2,\ldots,9\}$. Set $n_{1}= \mathrm{round}(\frac{n}{c_{0}+1})$
as the number of nodes in cluster 1 where $\mathrm{round}(x)$ denotes the nearest integer for any real number $x$. Note that $c_{0}$ is the ratio \footnote{Number of nodes in cluster 2 is $n-\mathrm{round}(\frac{n}{c_{0}+1})\approx n-\frac{n}{c_{0}+1}=c_{0}\frac{n}{c_{0}+1}$, therefore
	number of nodes in cluster 2 is around $c_{0}$ times of that in cluster 1.} of the sizes of
cluster 2 and cluster 1. We generate $\ell$ such that for $i=1,\dots,n_{1}, ~g_{i}=1$;
for $i=(n_{1}+1),\dots,n, ~g_{i}=2$.  And the mixing matrix $P_{2(d)}$ is as follows:
  \[\renewcommand{\arraystretch}{0.75}
P_{2(d)}
=
\begin{bmatrix}
    0.9&0.5\\
    0.5&0.9
\end{bmatrix}.
\]
Let $\theta$ be $\theta(i)=0.4$ if $g_{i}=1$ and $\theta(i)=0.6$ otherwise.

\emph{Experiment 2(e)}. All parameters are same as Experiment 2(d), except that $\theta_{i}=0.5+0.5(i/n)^{2}$ for $1\leq i\leq n$ (i.e., Experiment 2(e) is the DCSBM case).

\emph{Experiment 2(f)}. All parameters are same as Experiment 2(d), except that $\theta_{i}=0.5+0.5i/n$ for $1\leq i\leq n$.

\begin{figure}[!t]
\includegraphics[width=12cm,height=8cm]{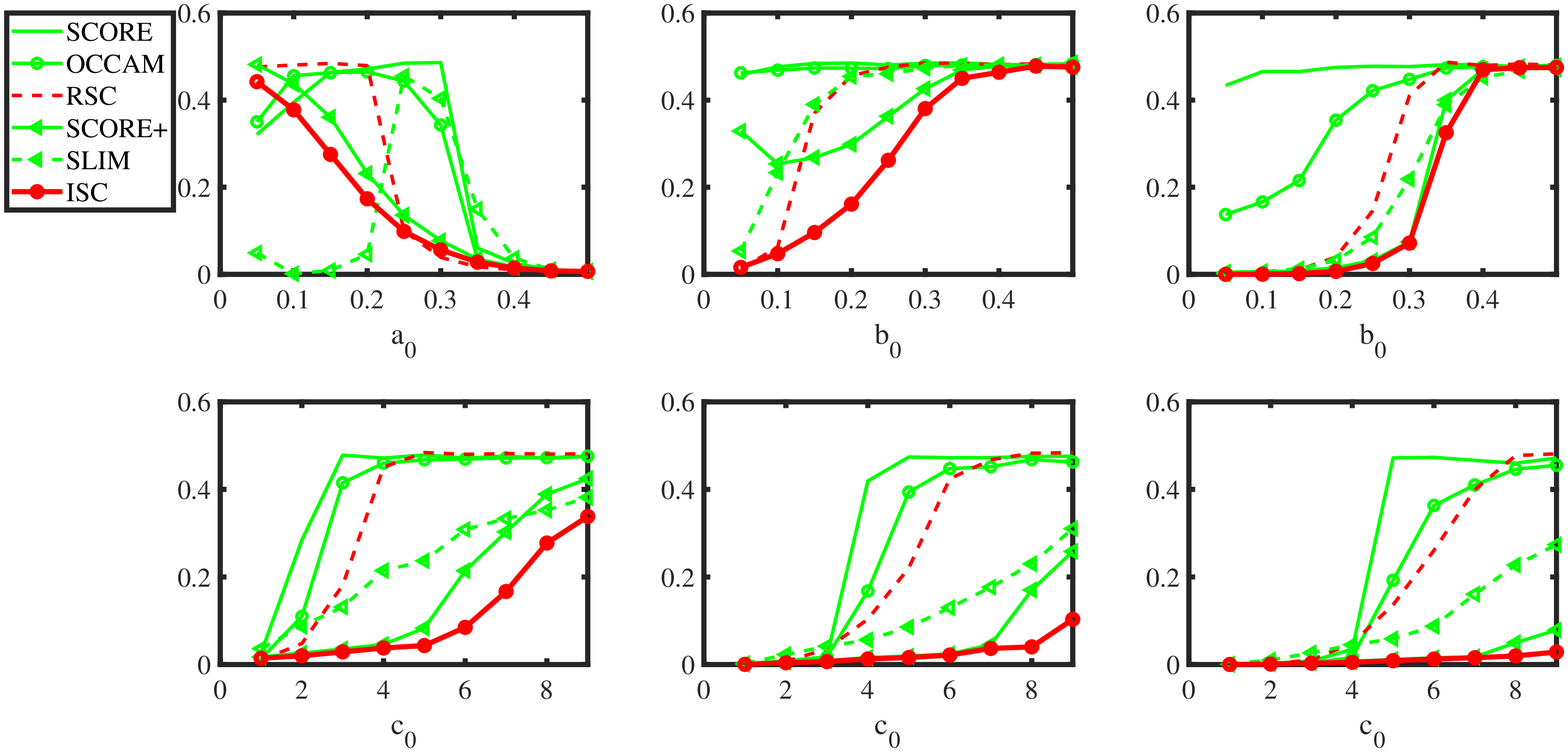}
\caption{Numerical results of Experiment 2. Top three panels (from left to right): Experiment 2(a), Experiment 2(b), and Experiment 2(c). Bottom three panels (from left to right): Experiment 2(d), Experiment 2(e), and Experiment 2(f). y-axis: error rates. }\label{Ex2}
\end{figure}

\begin{figure}[!t]
\includegraphics[width=12cm,height=8cm]{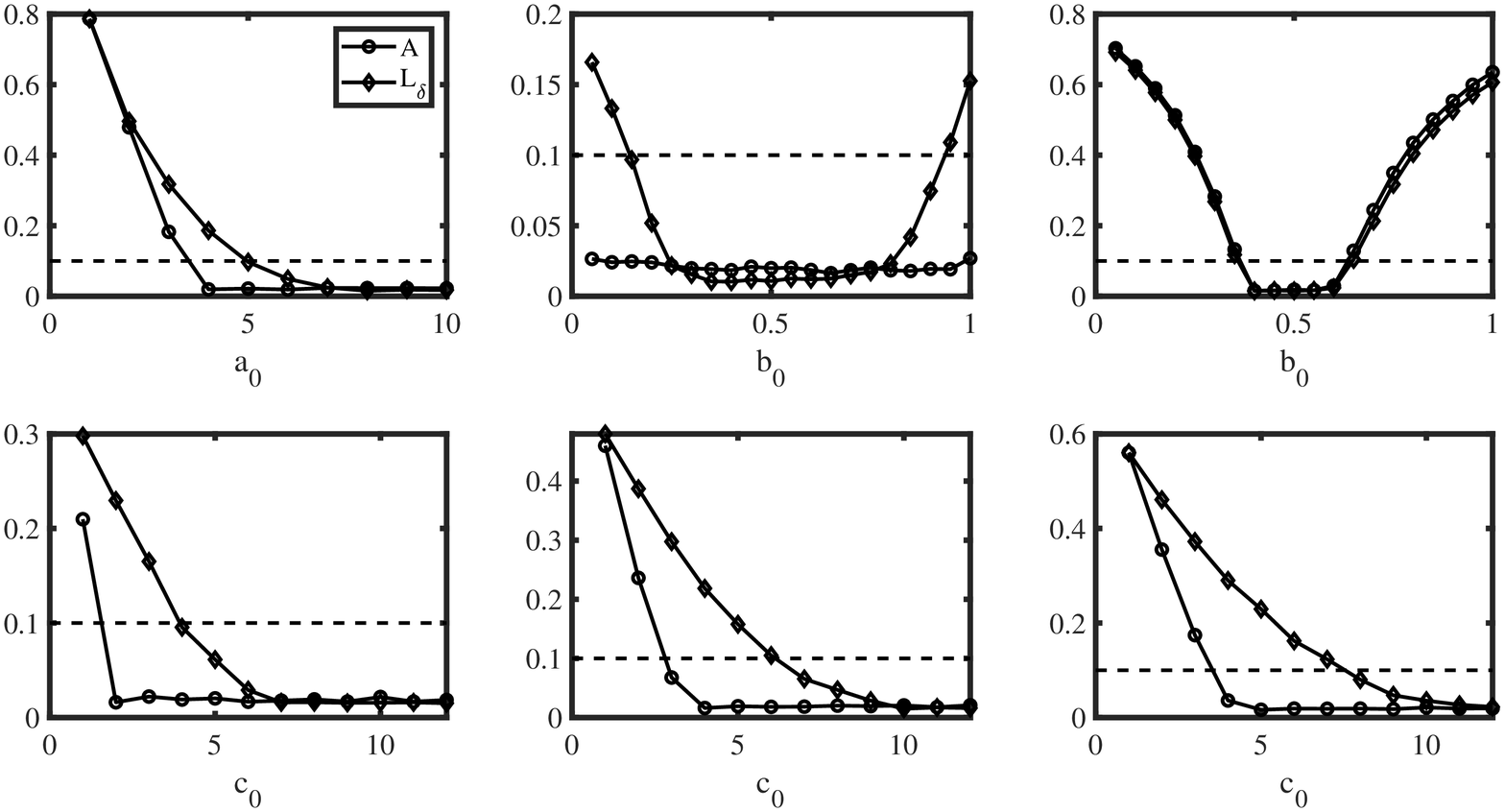}
\caption{Numerical results of Experiment 2. Top three panels (from left to right): Experiment 2(a), Experiment 2(b), and Experiment 2(c). Bottom three panels (from left to right): Experiment 2(d), Experiment 2(e), and Experiment 2(f). y-axis: $1-|\frac{\lambda_{K+1}}{\lambda_{K}}|$.}\label{eigEx2}
\end{figure}

The numerical results of Experiment 2 are demonstrated in Figure \ref{Ex2}, from which we can find that 1) ISC always have better performances than the other five approaches in this experiment. 2) numerical results of Experiment 2(d), 2(e), and 2(f) tells us that though it becomes challenging for all the six approaches to have satisfactory detection performances for a fixed size network when the size of one of the cluster decreases, our approach ISC always outperforms other five approaches. Recall that when the quantity $1-|\frac{\lambda_{K+1}}{\lambda_{K}}|$ is larger than 0.1, the network is a strong signal network. Figure \ref{eigEx2} shows that we generate both strong and weak signal networks in this experiment. When the network is weak signal, our ISC always performs satisfactory shown by Figure \ref{Ex2}. Especially, in the last three panels of Figure \ref{Ex2}, our ISC performs much better than other methods when the networks are strong signal. Therefore, generally speaking, numerical results of Experiment 2 concludes that our ISC almost always outperforms the other five procedures whether the simulated network is strong signal or weak signal.
Combining the numerical results of the above two experiments, we can draw a conclusion that our ISC has significant advantages over the three classical spectral clustering procedures SCORE, OCCAM and RSC, especially when dealing with weak signal networks. Numerical results also show that our ISC outperforms the recent spectral clustering approach SLIM. Though SCORE+ in \cite{SCORE+} is designed based on applying one more eigenvector for clustering when dealing with weak signal networks, our ISC outperforms SCORE+ in Experiment 1 and Experiment 2. This statement is also supported by the results of the next sub-section where we deal with eight real-world networks (two of them Simmons and Caltech are typical weak signal networks). Meanwhile, by observing the quantity $1-|\frac{\lambda_{K+1}}{\lambda_{K}}|$ where the two eigenvalues are from $A$ or $L_{\delta}$ in the above two experiments, we find that the networks we generated are weak signal networks numerically. The generation of weak signal networks is important since it can help us as well as readers to test the designed algorithms whether they can detect information of nodes labels for weak signal networks.
\subsection{Application to real-world datasets}
In this paper, eight real-world network datasets are analyzed to test the performances of our ISC.
The eight datasets are used in \cite{SCORE+} and can be downloaded directly from
\url{http://zke.fas.harvard.edu/software.html}. Table \ref{real8} presents some basic information about the eight datasets. These eight datasets are networks with known labels for all nodes where the true label information is surveyed by researchers. From Table \ref{real8}, we can see that $d_{\mathrm{min}}$ and $d_{\mathrm{max}}$ are always quite different for any one of the eight real-world datasets, which suggests a DCSBM case.  The true labels of the eight real-world datasets are originally suggested by the
authors/creators, and we take them as the ``ground truth''. Note that just as \cite{SCORE+}, there is pre-processing of the eight datasets because some nodes may have mixed memberships. For the Polbooks data, the books labeled as ``neutral'' are removed. For the football network, the five ``independent'' teams are deleted. For the UKfaculty data, the smallest group with only 2 nodes is removed. Therefore, after such pre-processing, the assumption that the given network consists of $K$ ``non-overlapping'' communities is reasonable and satisfied by the eight real-world networks.   Readers who are interested in the background information of the eight real-world networks can refer to \cite{SCORE+} or the source papers listed in Table \ref{real8} for more details.

\begin{table}
\footnotesize
\centering
\caption{Eight real-world data sets with known label information analyzed in this paper.}
\label{real8}
\resizebox{\columnwidth}{!}{
	\begin{tabular}{cccccccccc}
		\toprule
		Dataset& Source& $n$ & $K$& $d_{\mathrm{min}}$&$d_{\mathrm{max}}$\\\hline
		Karate& \cite{zachary1977an}&34&2&1&17\\
		Dolphins&\cite{lusseau2003the, lusseau2007evidence, lusseau2003theD}&62&2&1&12\\
		Football&\cite{girvan2002community}&110&11&7&13\\
		Polbooks&\cite{SCORE+}&92&2&1&24\\
		UKfaculty&\cite{nepusz2008fuzzy}&79&3&2&39\\
		Polblogs&\cite{adamic2005the}&1222&2&1&351\\
		Simmons&\cite{traud2011comparing}&1137&4&1&293\\
		Caltech&\cite{traud2011comparing}&590&8&1&179\\\hline
\end{tabular}}
\end{table}

Table \ref{eigratio} presents the quantity of $1-|\frac{\lambda_{K+1}}{\lambda_{K}}|$ for the eight real-world datasets. It needs to mention that when $1-|\frac{\lambda_{K+1}}{\lambda_{K}}|<0.1$, the $(K+1)$-th eigenvalue is quite close to the $K$-th one. From Table \ref{eigratio} we can find that the quantities for $A$ of Karate, Football, Simmons and Caltech are smaller than 0.1. While the quantities for $L_{\delta}$ are larger than 0.1 for Karate and Football but smaller than 0.1 for Simmons and Caltech. As discussed in \cite{SCORE+}, we know Karate and Football are usually taken as strong signal networks, while Simmons and Caltech are weak signal networks. Therefore, the quantity of $L_{\delta}$ could be more effective to distinguish weak and strong signal networks.
\begin{table}
\centering
\caption{The quantity $1-|\frac{\lambda_{K+1}}{\lambda_{K}}|$ for the eight real-world datasets, where $\lambda_{k}$ are the $k$-th leading eigenvalues from $A$ or $L_{\delta}$.}
\label{eigratio}
\resizebox{\linewidth}{!}{\begin{tabular}{cccccccccc}
\toprule
$1-|\frac{\lambda_{K+1}}{\lambda_{K}}|$ &Karate&Dolphins&Football&Polbooks&UKfaculty&Polblogs&Simmons&Caltech\\
\midrule
$A$&0.0984&0.1863&0.0213&0.5034&0.2996&0.5101&0.0804&0.0777\\
$L_{\delta}$&0.1610&0.2116&0.1469&0.2720&0.3666&0.4570&0.0540&0.0241\\
\bottomrule
\end{tabular}}
\end{table}

\begin{table}[t!]
\centering
\caption{Error rates on the eight empirical data sets.}
\label{real8errors}
\resizebox{\linewidth}{!}{\begin{tabular}{cccccccccc}
\toprule
\textbf{ Methods} &Karate&Dolphins&Football&Polbooks&UKfaculty&Polblogs&Simmons&Caltech\\
\midrule
SCORE&\textbf{0/34}&\textbf{0/62}&5/110&\textbf{1/92}&1/79&58/1222&268/1137&180/590\\
$\mathrm{SCORE}_{K+1}$&\textbf{0/34}&\textbf{0/62}&5/110&24/92&35/79&281/1222&187/1137&150/590\\
OCCAM&\textbf{0/34}&1/62&4/110&3/92&5/79&60/1222&268/1137&192/590\\
RSC&\textbf{0/34}&1/62&5/110&3/92&\textbf{0/79}&64/1222&244/1137&170/590\\
$\mathrm{RSC}_{K+1}$&\textbf{0/34}&15/62&4/110&3/92&2/79&79/1222&134/1137&100/590\\
SCORE+&1/34&2/62&6/110&2/92&2/79&\textbf{51/1222}&127/1137&98/590\\
SLIM&1/34&\textbf{0/62}&6/110&2/92&1/79&\textbf{51/1222}&275/1137&150/590\\
$\mathrm{SLIM}_{K+1}$&1/34&\textbf{0/62}&6/110&2/92&2/79&53/1222&268/1137&157/590\\
CMM&\textbf{0/34}&1/62&7/110&\textbf{1/92}&7/79&62/1222&137/1137&106/590\\
\hline
ISC&\textbf{0/34}&1/62&\textbf{3/110}&3/92&1/79&64/1222&\textbf{121/1137}&\textbf{96/590}\\
\bottomrule
\end{tabular}}
\end{table}
Next we study the performances of these methods (note that we also compare our ISC with the convexied modularity maximization (CMM) method by \cite{CMM}) on the eight real-world networks. First, we'd note that in the procedure of SCORE, RSC and SLIM, there is one common step which computes the leading $K$ eigenvectors of $A$ or its variants. For fair comparison, we set $\mathrm{SCORE}_{K+1}$, $\mathrm{RSC}_{K+1}$ and $\mathrm{SLIM}_{K+1}$ as three new algorithms such that $\mathrm{SCORE}_{K+1}$ is SCORE method but using the leading $K+1$ eigenvectors, $\mathrm{RSC}_{K+1}$ is RSC by applying the leading $K+1$ eigenvectors, and $\mathrm{SLIM}_{K+1}$ is SLIM by applying the leading $K+1$ eigenvectors (i.e., ISC, $\mathrm{SCORE}_{K+1}$, $\mathrm{RSC}_{K+1}$, and $\mathrm{SLIM}_{K+1}$ all apply $K+1$ eigenvectors for clustering).

Table \ref{real8errors} summaries the error rates on the eight real-world networks. For Karate, Dolphins, Football, Polboks, UKfaculty, and Polblogs, ISC has similar performances as SCORE, OCCAM, RSC and CMM, while SCORE+ and SLIM perform best with the error rate 51/1222 for Polblogs. However, for the Football network, ISC has the smallest number of errors while CMM has the largest. Though $\mathrm{SCORE}_{K+1}$, $\mathrm{RSC}_{K+1}$ and $\mathrm{SLIM}_{K+1}$ also apply the leading $K+1$ eigenvectors for clustering, these three approaches fail to detect some of the eight real-world datasets with pretty high error rates. For instance, $\mathrm{SCORE}_{K+1}$ fails to detect Polbooks, Ukfaculty and Polblogs, meanwhile $\mathrm{RSC}_{K+1}$ fails to detect Dolphins.  When dealing with Simmons and Caltech, ISC has excellent performances on these two datasets and significantly outperforms all other approaches with 121/1137 error rate for Simmons and 96/590 for Caltech. $\mathrm{SLIM}_{K+1}$ perform similar as SLIM and both two procedures fail to detect Simmons and Caltech with high error rates.  Deserved to be mentioned, $\mathrm{SCORE}_{K+1}$ and $\mathrm{RSC}_{K+1}$ perform better than the original SCORE and RSC for Simmons and Caltech, respectively. This phenomenon occurs because the leading $K+1$ eigenvalue of $A$ or its variants for Simmons (Caltech) is close to the leading $K$ eigenvalue as shown in the Table \ref{eigratio}, and hence the leading $K+1$ eigenvector also contains label information.  


Then we study the different choice of the tuning parameters $\delta$. In Table \ref{real8errordelta}, set $\delta \in \{0, 0.025, 0.05, \ldots, 0.175, 0.20\}$, and fix other parameters and record the corresponding number of errors for the eight real-world networks. The results of Table \ref{real8errordelta} tells us that ISC successfully detect these networks except that it fails to detect Simmons when $\delta$ is 0 or 0.025. Since when $\delta$ is set as 0, $L_{\delta}$ is the Laplacian matrix $L$. Therefore, according to the results of Table \ref{real8errordelta}, we suggest that $\delta$ should be larger than 0. Next, we further study the choice of $\delta$ by taking values in $\{0.005, 0.01, 0.015 \ldots, 1\}$ (there are 200 choices of $\delta$) and $\delta\in\{1, 2, \ldots, 200\}$ (there are 200 choices of $\delta$). The number of errors for the eight real-world networks with a variety of $\delta$ are plotted in the Figure \ref{delta}. From Figure \ref{delta} we can see that ISC always successfully detect information of nodes labels for the eight real-world networks even when $\delta$ is set as large as hundreds or as small as 0.005. The results of Table \ref{real8errordelta} and Figure \ref{delta} suggest that ISC is insensitive to different choices of $\delta$ as long as $\delta>0$, and we can even set $\delta$ as large as hundreds. Actually, this phenomenon is supported by Theorem \ref{mainISC}, which tell us that when fixing parameters $n, P, \Theta, Z$ under DCSBM and increasing $\delta$, two assumptions (a) and (b) in Lemma \ref{boundstar} still hold, which suggests the feasibility of our ISC.

Finally, we study the value $d=\frac{d_{\mathrm{max}}+d_{\mathrm{min}}}{2}$. Note that the value $\delta d$ in our regularized Laplacian matrix $L_{\delta}$ ($L_{\delta}=(D+\delta d I)^{-1/2}A(D+\delta d I)^{-1/2}$), actually, is the regularizer $\tau$ in the regularized Laplacian matrix $L_{\tau}$ ($L_{\tau}=(D+\tau I)^{-1/2}A(D+\tau I)^{-1/2}$) which is defined in the RSC approach and the default choice of $\tau$ is set as the average degree. Then, one question arises naturally, whether we can apply other values such as $d_{\mathrm{max}}, d_{\mathrm{min}}$ or $\bar{d}$ (the average degree) to replace $d$ in our ISC method? The answer is YES. We replace $d$ by $d_{\mathrm{max}}, d_{\mathrm{min}}$ and $\bar{d}$ in our algorithm, and demonstrate the numerical results in Table \ref{real8errord} with notation $\mathrm{ISC}_{d_{\mathrm{max}}}, \mathrm{ISC}_{d_{\mathrm{min}}}$ and $\mathrm{ISC}_{\bar{d}}$, respectively.  From Table \ref{real8errord}, we can find that 1) $\mathrm{ISC}_{d_{\mathrm{min}}}$ fails to detect Simmons and Caltech with number of errors as large as 305 and 154, respectively; 2)  $\mathrm{ISC}_{\bar{d}}$ fails to detect Simmons with number of errors as large as 200; 3) $\mathrm{ISC}_{d_{\mathrm{max}}}$ performs better than $\mathrm{ISC}_{d_{\mathrm{min}}}$ and $\mathrm{ISC}_{\bar{d}}$, and ISC outperforms $\mathrm{ISC}_{d_{\mathrm{max}}}$ for all datasets. As shown by the numerical results in Table \ref{real8errord}, ISC almost always performs better than $\mathrm{ISC}_{d_{\mathrm{max}}}$, $\mathrm{ISC}_{d_{\mathrm{min}}}$ and $\mathrm{ISC}_{\bar{d}}$, which suggests us the default choice of $d$ is $d=\frac{d_{\mathrm{max}}+d_{\mathrm{min}}}{2}$ in this paper.
\begin{rem}
	According to the numerical results with several $d$ in Table \ref{real8errord}, we find that ISC is sensitive to the choice of $d$ since $\mathrm{ISC}_{d_{\mathrm{min}}}$ fails to detect Simmons and Caltech, $\mathrm{ISC}_{\bar{d}}$ fails to detect Simmons. Luckily, though ISC is sensitive to the choice of $d$, when $d$ is set as $d=\frac{d_{\mathrm{max}}+d_{\mathrm{min}}}{2}$, ISC always has satisfactory performances according to all the numerical results in this paper. We argue that whether there exists an optimal $d$ (instead of simply setting it as $\frac{d_{\mathrm{max}}+d_{\mathrm{min}}}{2}$ based on the numerical results) based on rigorous theoretical analysis, and we leave it as future work.
\begin{table}[t!]
\centering
\caption{Community detection errors of ISC on the eight empirical data sets for different $\delta$.}
\label{real8errordelta}
\resizebox{\linewidth}{!}{\begin{tabular}{cccccccccc}
\toprule
$\delta$ &Karate&Dolphins&Football&Polbooks&UKfculty&Polblogs&Simmons&Caltech\\
\midrule
0&1&1&3&3&2&60&307&122\\
0.025&1&1&3&3&2&61&200&96\\
0.05&0&1&3&3&2&63&122&98\\
0.075&0&1&6&3&1&63&121&96\\
0.10&0&1&3&3&1&64&121&96\\
0.125&0&0&3&3&1&64&121&96\\
0.15&0&0&3&3&1&65&121&97\\
0.175&0&0&3&3&1&66&121&96\\
0.20&0&0&5&3&1&67&123&98\\
\bottomrule
\end{tabular}}
\end{table}
\begin{figure}[t!]
\includegraphics[width=12cm,height=10cm]{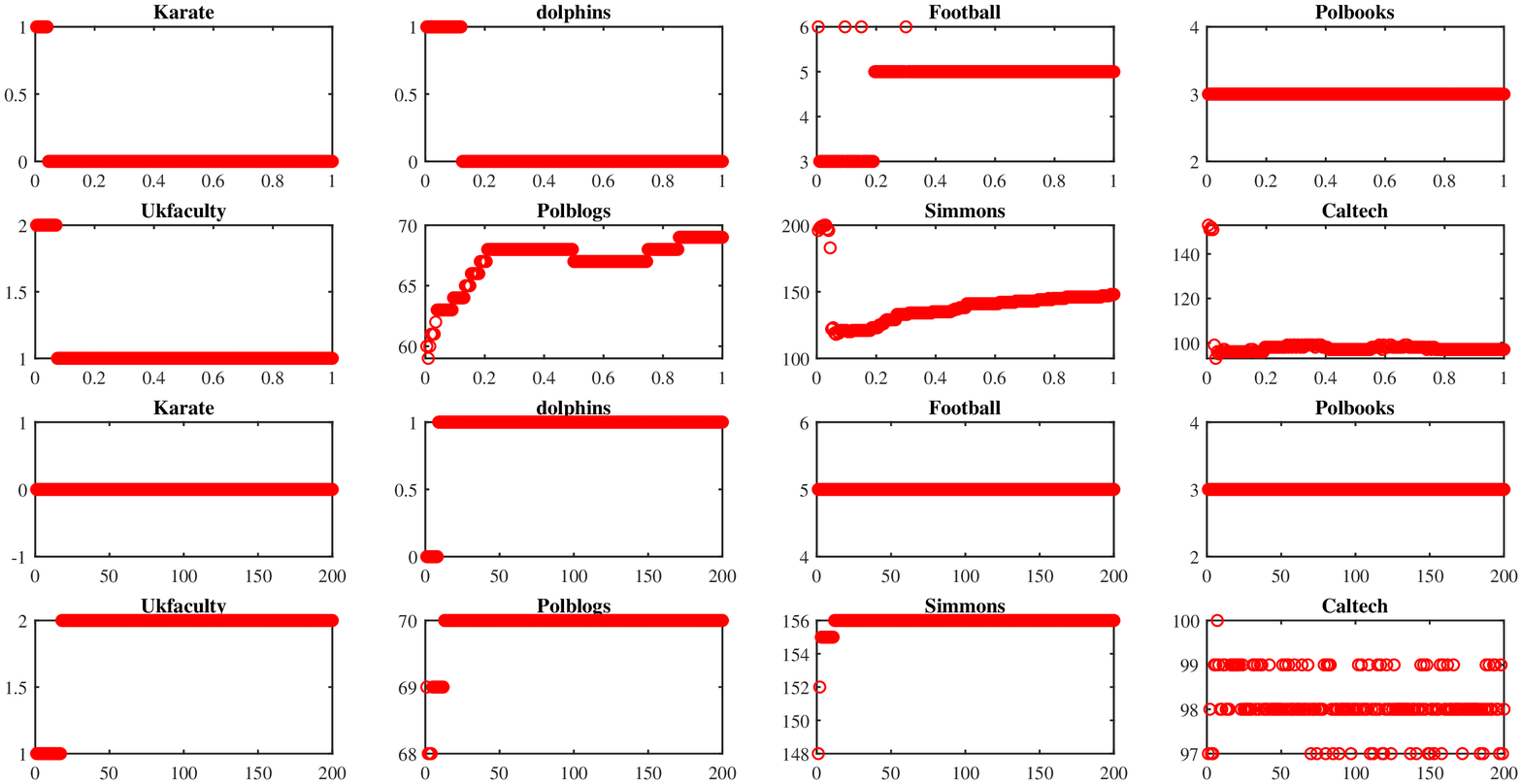}
\caption{Community detection errors of ISC on the eight empirical data sets for different $\delta$. The first eight panels: $\delta\in\{0.005, 0.01, 0.015 \ldots, 1\}$; the last eight panels: $\delta\in\{1, 2, \ldots, 200\}$. y-axis: number errors.}\label{delta}
\end{figure}
\begin{table}[t!]
\centering
\caption{Error rates on the eight empirical datasets.}
\label{real8errord}
\resizebox{\linewidth}{!}{\begin{tabular}{cccccccccc}
\toprule
\textbf{ Methods} &Karate&Dolphins&Football&Polbooks&UKfaculty&Polblogs&Simmons&Caltech\\
\midrule
ISC&0/34&1/62&3/110&3/92&1/79&64/1222&121/1137&96/590\\
$\mathrm{ISC}_{d_{\mathrm{max}}}$&0/34&0/62&3/110&3/92&1/79&67/1222&123/1137&98/590\\
$\mathrm{ISC}_{d_{\mathrm{min}}}$&1/34&1/62&6/110&3/92&2/79&60/1222&305/1137&154/590\\
$\mathrm{ISC}_{\bar{d}}$&0/34&1/62&3/110&3/92&2/79&60/1222&200/1137&96/590\\
\bottomrule
\end{tabular}}
\end{table}
\end{rem}
\section{Discussion}\label{sec5}
In this paper, we introduced a community detection method named ISC based on the production of the leading eigenvalues and eigenvectors of the regularized graph Laplacian in networks. Under DCSBM, we established theoretical proprieties for the proposed method. From the simulation and empirical results we can find that the ISC method demonstrably outperforms classical spectral clustering methods (such as SCORE, RSC, OCCAM and SCORE+) and the newly-published community detection method  SLIM, especially for weak signal networks.
The most important step in our proposed method is that we always apply $K+1$ eigenvectors for clustering. Thus we implemented this idea on other competitors, SCORE, RSC and SLIM, which are denoted as $\mathrm{SCORE}_{K+1}$, $\mathrm{RSC}_{K+1}$ and $\mathrm{SLIM}_{K+1}$ in numerical studies. Meanwhile, as demonstrated in the numerical results of eight real-world networks, $\mathrm{SCORE}_{K+1}$ and $\mathrm{SLIM}_{K+1}$ fail to detect some real-world networks, such as Dolphins, Football, UKfaculty and Polblogs while $\mathrm{SLIM}_{K+1}$ performs poor on the two weak signal networks  Simmons and Caltech. As for OCCAM, it can not consider $K+1$ eigenvectors for clustering. By \cite{OCCAM}, we know that OCCAM is designed originally for overlapping community detection problems, therefore it's meaningful to extend OCCAM for weak signal networks, and we leave it as future work. Meanwhile, since both $\mathrm{SLIM}_{K+1}$ and SLIM have unsatisfactory performances on Simmons and Caltech, we may conclude that it is challenging to extend the spectral clustering approaches based on the symmetric Laplacian inverse matrix (SLIM) \cite{SLIM} to function satisfactory on weak signal networks, and we leave it as future work. Overall, our method ISC is appealing and indispensible in the area of community detection, especially when dealing with weak signal networks.


There are several open problems which drive us to some interesting and meaningful extensions of our ISC method. For example, it remains unclear about how to build the theoretic framework for the generation of weak signal networks, such as directed, weighted, dynamic and mixed membership networks, etc. Therefore, extending ISC to the above weak signal networks is significant. It also remains unclear that whether there exist optimal $\delta$ and $d$ for ISC.  Meanwhile, constructing newly variants of the adjacency matrix $A$ may help us to design spectral method with even better performances than ISC for weak signal networks. Recall that we always assume that the number of communities $K$ is known in advance for a given network and aim to detect nodes labels. However, in practice, the true number of clusters is usually unknown, and when we turn to weak signal networks, because the $(K+1)$-th eigenvalue is close to the $K$-th one of $A$ or its variants, traditional methods (\cite{chen2018network, hu2019corrected, le2015estimating}) aiming at estimating the number of clusters may fail to find the true number of clusters of weak signal networks. So it is crucial and meaningful for researchers to construct methods to estimate the number of clusters for both strong signal and weak signal networks. Based on the special form of the OCCAM algorithm, we know that it can not apply $(K+1)$ eigenvectors for clustering to deal with weak signal networks, therefore it's meaningful to extend OCCAM to solve weak signal networks' community detection problem. Furthermore, more than the mixed membership community detection, the question that dealing with dynamic weak signal networks may possibly occur in the multiple networks community detection problem as in \cite{arroyo2020simultaneous}. Finally, since both $\mathrm{SLIM}_{K+1}$ and SLIM fail to detect Simmons and Caltech, it is meaningful to design spectral clustering algorithms based on the symmetrized Laplacian inverse matrix \cite{SLIM} for both strong signal and weak signal networks.  We leave studies of these problems to our future work.
\bibliographystyle{Chicago}
\bibliography{refISC}

\end{document}